\documentclass[letterpaper, 10 pt, conference]{ieeeconf}  

\IEEEoverridecommandlockouts                              

\usepackage[OT1]{fontenc} 
\usepackage{graphicx} 
\usepackage{multirow}
\usepackage{colortbl}
\usepackage{algorithm}
\usepackage{algpseudocode}
\usepackage{amsmath}
\usepackage{amsfonts}
\usepackage{stfloats}
\usepackage{subfigure}
\usepackage{makecell}
\usepackage{float}
\usepackage{caption}
\usepackage{threeparttable}
\usepackage[hyperref=true,style=ieee,backend=biber,sorting=none,maxnames=1,minnames=1]{biblatex}
\makeatletter
\let\NAT@parse\undefined
\makeatother
\usepackage[colorlinks,linkcolor=blue,anchorcolor=blue]{hyperref}
\title{\LARGE \bf
	Fine-Grained Off-Road Semantic Segmentation and Mapping via Contrastive Learning
}

\author{Biao Gao$^{1}$, Shaochi Hu$^{1}$, Xijun Zhao$^{2}$, Huijing Zhao$^{1}$
	\thanks{*This work is supported in part by the National Natural Science Foundation of China under Grant 61973004.}
	\thanks{$^{1}$B. Gao, S. Hu and H. Zhao are with the Key Lab of Machine Perception (MOE), Peking University, Beijing, China. $^{2}$X. Zhao is with China North Vehicle Research Institute, Beijing, China.}%
	\thanks{Correspondence: H. Zhao, {\tt\small zhaohj@cis.pku.edu.cn}.}%
}

\begin{document}
	\let\oldtwocolumn\twocolumn
	\renewcommand\twocolumn[1][]{%
		\oldtwocolumn[{#1}{
			\begin{center}
				\vspace{-3mm}
				\includegraphics[width=0.95\textwidth]{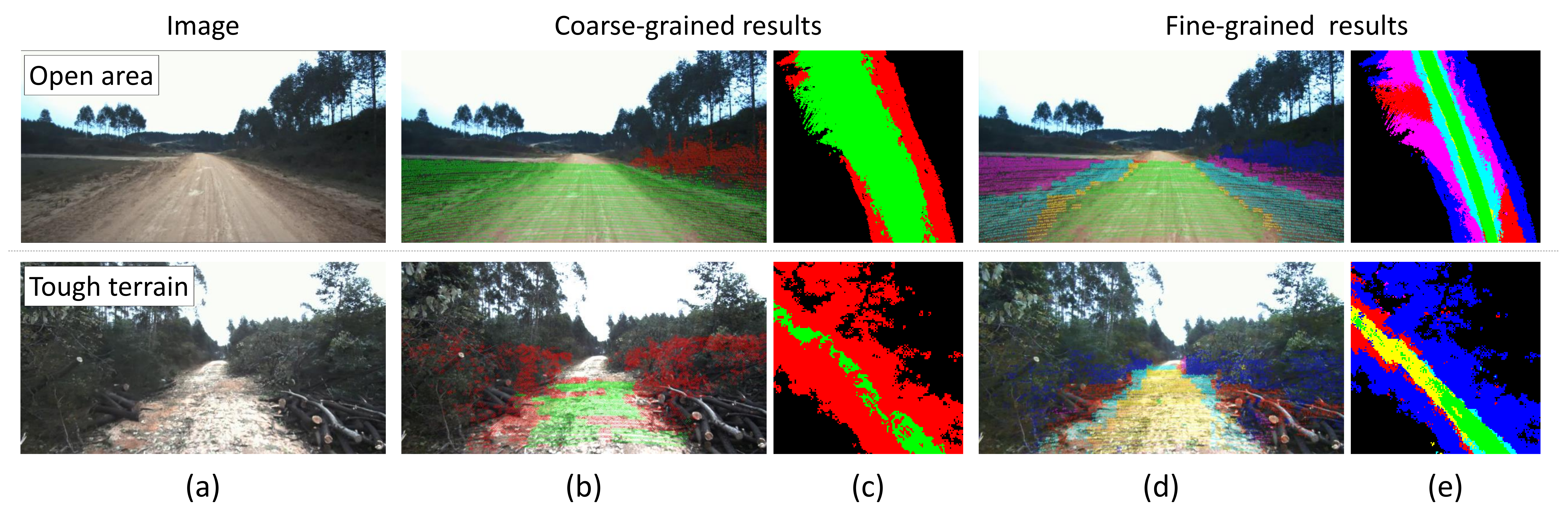}
				\vspace{-2mm}
				\captionof{figure}{The significance of fine-grained semantic segmentation and mapping in off-road environment, where coarse-grained results can hardly adapt diverse scenes. (a) scene image. (b) coarse-grained semantic segmentation (binary classification). (c) coarse-grained semantic map. (d) fine-grained semantic segmentation. (e) fine-grained semantic map.}
				\label{fig:problem_def}
			\end{center}
		}]
	}
	
	\maketitle

	
\begin{abstract}
Road detection or traversability analysis has been a key technique for a mobile robot to traverse complex off-road scenes.
The problem has been mainly formulated in early works as a binary classification one, e.g. associating pixels with road or non-road labels.
Whereas understanding scenes with fine-grained labels are needed for off-road robots, 
as scenes are very diverse, and the various mechanical performance of off-road robots may lead to different definitions of safe regions to traverse.
How to define and annotate fine-grained labels to achieve meaningful scene understanding for a robot to
traverse off-road is still an open question.
This research proposes a contrastive learning based method. 
With a set of human-annotated anchor patches, a feature representation is learned to discriminate regions with different traversability, 
a method of fine-grained semantic segmentation and mapping is subsequently developed for off-road scene understanding.
Experiments are conducted on a dataset of three driving segments that represent very diverse off-road scenes.
An \textit{anchor accuracy} of 89.8\% is achieved by evaluating the matching with human-annotated image patches in cross-scene validation.
Examined by associated 3D LiDAR data, the fine-grained segments of visual images are demonstrated to have different levels of toughness and terrain elevation, which represents their semantical meaningfulness.
The resultant maps contain both fine-grained labels and confidence values, providing rich information to support a robot traversing complex off-road scenes.

\end{abstract}
	
	\section{Introduction}
	Mobile robotic and autonomous driving techniques have been witnessed with tremendous progress in recent years~\cite{feng2020deep}. Driving scene understanding plays a vital role as a prerequisite for the decision making and planning of a robot to traverse in complex environments~\cite{badue2020self}.
	Nowadays researches are mainly oriented to the applications at structural scenes such as indoor, parking lots, urban streets, highways, etc.~\cite{siam2017deep},
	whereas research on understanding off-road environments is rare.
	The off-road environment is unstructured, dominated by natural objects, lacking artificial features, and its terrain conditions are various and complex.
	One of the fundamental techniques of an off-road robot is to detect safe regions (hereinafter called {\it off-road}) to traverse, which has also been termed as traversable surface~\cite{zhou2012self}, drivable corridor~\cite{nefian2006detection}, etc., in literature.
	Comparing with the roads in structured environments, where functional attributes are clearly defined by artificial features such as pavement, barrier, and markings,
	off-roads are ill-defined~\cite{ososinski2015automatic}. 
	
	Early methods of off-road detection are usually developed by assuming color, texture, boundaries of the target, where rule-based methods of extracting vanishing point and subsequently road boundaries~\cite{kong2009vanishing}\cite{shi2015fast}, and segmentation-based methods of extracting continuous regions based on certain road models are developed~\cite{alon2006off}\cite{wang2009unstructured}.
	These methods are called {\it coarse-grained} ones as the problem is formulated as a binary classification, e.g. labeling each image pixel to {\it road} or {\it non-road}.
	As illustrated in Fig. \ref{fig:problem_def}(b-c), such methods may fail to detect any region to traverse at tough terrains or extract too wide regions that lack efficiency in promoting the best choice at open area.
	Moreover, the mechanical performance of off-road robots can be very different, leading to different definitions and selections of safe regions to traverse.
	Understanding scenes with fine-grained labels is needed for off-road robots~\cite{wellhausen2019should}.
	On the other hand, deep learning methods have been studied in recent years~\cite{rateke2019passive}. 
	Semantic segmentation using deep learning techniques infers scenes at pixel- or point-levels~\cite{long2015fully}, where large-scale datasets such as Cityscapes~\cite{cordts2016cityscapes}, SemanticKITTI~\cite{behley2019semantickitti} with fine-grained labels and massive annotations are needed.
	There is no such dataset at off-road scenes. How to define and annotate fine-grained labels to achieve meaningful scene understanding for a robot to traverse off-road is still an open question.  
	
	This research proposes a contrastive learning method to achieve fine-grained semantic segmentation and mapping of off-road scenes as shown in Fig. \ref{fig:problem_def}(d-e).
	It is difficult to define fine-grained categories that are generalized at diverse off-road scenes and it is further hard for a human operator to assign fine-grained labels to each image pixel, where the definitions could be very ambiguous at natural scenes. However, it is not difficult for a human operator to annotate images by sparse anchor patches as illustrated in Fig. \ref{fig:pipeline} to indicate the regions with different semantic attributes on their traversability. Inspired by impressive progress and promising results of contrastive learning~\cite{oord2018CPC}\cite{chen2020simple}\cite{he2020momentum}, this research learns a feature representation to discriminate regions with different semantic attributes using contrastive learning, which is used to develop a method of fine-grained semantic segmentation and mapping for off-road applications.
	An off-road dataset is developed containing over 12000 image frames of three driving segments that represent very diverse off-road scenes.
	With no more than 100 training frames in all experimental settings, the test results in cross-scene validation show an 89.8\% \textit{anchor accuracy}, which is a new metric introduced to evaluate the matching with human-annotated image patches. 
	Examined by additionally measured 3D LiDAR data, 
	it is found that the fine-grained segments of visual images are semantically meaningful to represent different levels of toughness and terrain elevation.
	The resultant maps contain both fine-grained labels and confidence values, providing rich information to support a robot traversing complex off-road environments.

	This paper is organized as follows. First, the related works
	are introduced in Section~\ref{related_works}. Section~\ref{methodology} presents the proposed methodology in detail. Section~\ref{exp} shows experimental results. Finally, we draw conclusions in Section~\ref{conclusions}.
	
	\section{Related Works} \label{related_works}
	\subsection{Rule/Segmentation-based Methods}
	Rule/segmentation-based methods are mainly developed by assuming color, texture, boundaries of the target region, and these researches are mostly coarse-grained understanding that formulates the problem as a binary classification. They can be broadly divided into rule-based and segmentation-based methods.
	
	Some rule-based methods utilize global priors like vanishing point~\cite{kong2009vanishing}\cite{shi2015fast}, which primarily depend on edge cues to obtain road area. The others assume the road region as geometric triangular~\cite{zhou2010self} or trapezoidal~\cite{jeong2002vision} shape.
	
	Segmentation-based methods formulate the problem as pixel-level segmentation tasks. Some studies~\cite{lu2014hierarchical}
	assume the region at bottom of images as road data or collect vehicle trajectories as drivable area~\cite{mei2017scene}, then label similar regions as roads.
	Other methods~\cite{alon2006off}\cite{wang2009unstructured} depend on fixed road models and make use of hybrid features to extract continuous regions.
	
	\subsection{Deep Learning Methods}
	Benefit by developments of deep networks~\cite{long2015fully} and large-scale datasets with fine-grained labels like Cityscapes~\cite{cordts2016cityscapes} and SemanticKITTI~\cite{behley2019semantickitti}, deep learning methods are able to get fine-grained semantic segmentation or maps. However, most existing datasets and studies are designed for urban scenes, and research in off-road environments is still limited.
	
	Due to the lack of datasets, studies for off-road scenes attempt several ways to reduce the demand for fine-annotated data, such as weakly and semi-supervised learning~\cite{suger2015traversability}\cite{gao2019off}, and transfer learning~\cite{holder2016road}\cite{sharma2019semantic}. 
	One mainstream idea is automatically generating training data from other sensor modalities, such as 3D LiDAR data~\cite{gao2019off}\cite{tang2017one}, audio features~\cite{zurn2020self} and force-torque signals~\cite{wellhausen2019should}.
	Another idea is to transfer knowledge of deep networks from existing urban datasets~\cite{holder2016road} or synthetic data~\cite{sharma2019semantic} to off-road environments.
	Nevertheless, transferred models still need some fine-annotated data for finetuning, and the performance is limited by domain gaps. Meanwhile, labels from other modalities or synthetic data are too limited to support fine-grained semantic segmentation and mapping.
	
	\subsection{Contrastive Learning}
	Recent progress in contrastive learning~\cite{oord2018CPC}\cite{chen2020simple}\cite{he2020momentum} demonstrates that discriminative representations could be learned through a self-supervised pipeline, by contrasting positive and negative samples. Various sample definitions make contrastive learning suitable for diverse domains like natural language~\cite{oord2018CPC} and images~\cite{tian2019contrastive}. 
	Zhao et al.~\cite{zhao2020contrastive} introduce contrastive learning to semantic segmentation task, but rely on pixel-level labeled data for initial contrastive learning and generating pseudo labels for unlabeled images.
	
	Inspired by the promising results of contrastive learning, but different from settings in~\cite{zhao2020contrastive}, this work only relies on a small number of sparse anchor annotations without pixel-level labels to learn feature representations to discriminate regions with different semantic attributes, which is further used to develop a method of fine-grained semantic segmentation and mapping.
	
	\begin{figure*}[]
		\centering
		\includegraphics[width=\textwidth]{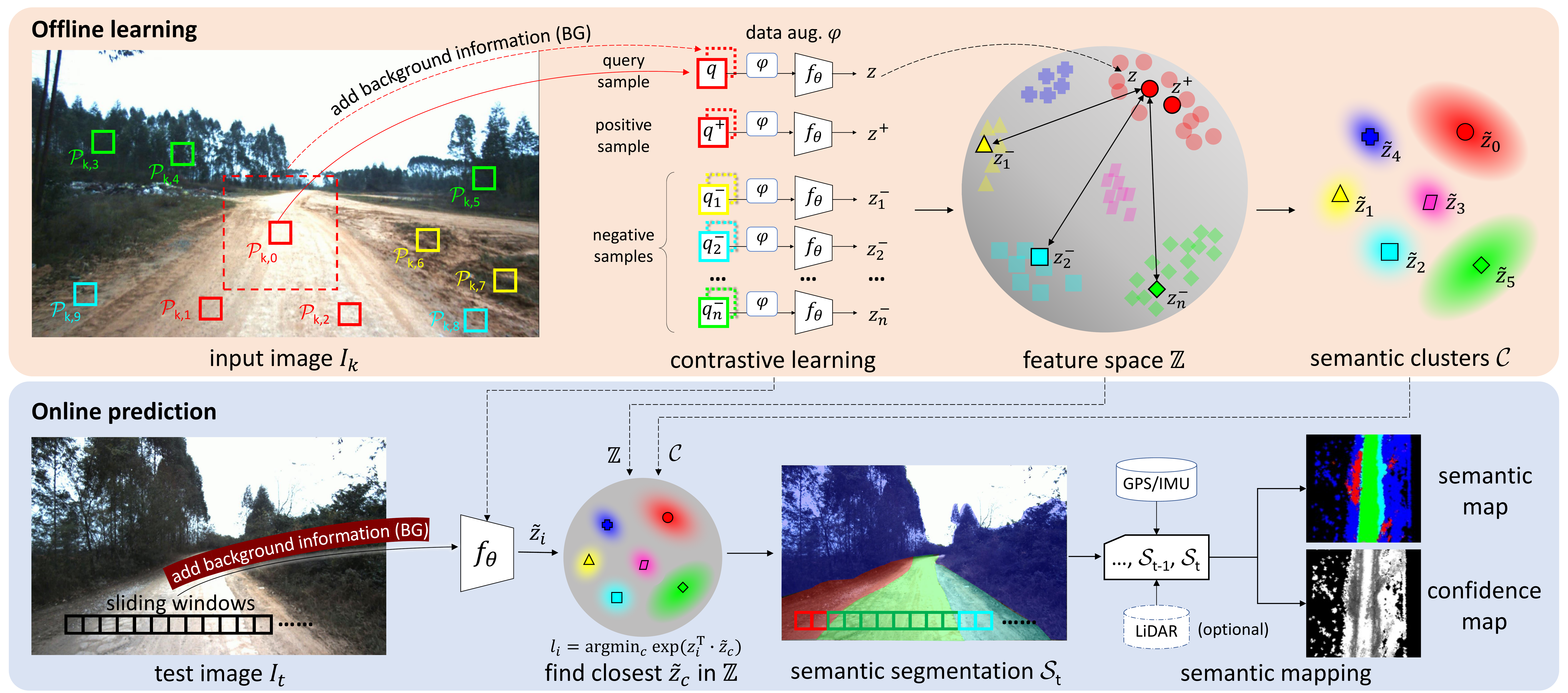}
		\caption{The proposed pipeline for fine-grained off-road semantic segmentation and mapping via contrastive learning.}
		\label{fig:pipeline}
		\vspace{-2mm}
	\end{figure*}
	
	\section{Methodology} \label{methodology}
	
	\subsection{Problem Formulation}
	
	A training image $I_k$ has a number of anchor patches $A_k=\{\mathcal{P}_{k,i}=<p_{k,i},a_{k,i}>\}$, where an anchor patch $\mathcal{P}_{k,i}$ is a pair of an image patch $p_{k,i}$ and a label $a_{k,i}$. Here, $a_{k,i}$ has no semantic meaning, but is an identifier of the image patches with similar or different semantic properties.
	Let $z=f_{\theta}(p)$ be an encoder converting a high-dimensional image patch $p$ to a normalized low-dimensional feature vector $z\in \mathbb{Z}^D$. 
	We use exponential cosine similarity $sim(p_i,p_j)=exp(z_i^T \cdot z_j)$ to measure the similarity of two image patches via their low-dimensional feature vectors.
	Therefore, given an anchor patch $\mathcal{P}_{k,i}$, its similarity to another anchor patch $\mathcal{P}_{k,j}$, i.e. $sim(p_{k,i},p_{k,j})$, should be higher if they share the same label $a_{k,i}=a_{k,j}$, whereas lower if the labels are different $a_{k,i} \neq a_{k,j}$.
	In order to make the annotation operational easy, in this research, the labels of the anchor patches are comparable only if they belong to the same image.
	
	Given a set of training images $\mathcal{I}=\{I_k\}$ with anchor patches $\mathcal{A}=\{A_k\}$ on each of them, this research is to find a representation $f_{\theta}$ that encodes image patch $p$ to $z$, where at the low-dimensional feature space $\mathbb{Z}^D$, the $z$ of similar semantic meaning distribute closely. 
	This research finds $f_{\theta}$ through contrastive learning, which is further used in an application of fine-grained semantic segmentation and mapping for off-road traversability analysis.
	
	\begin{figure}[]
		\centering
		\includegraphics[scale=0.23]{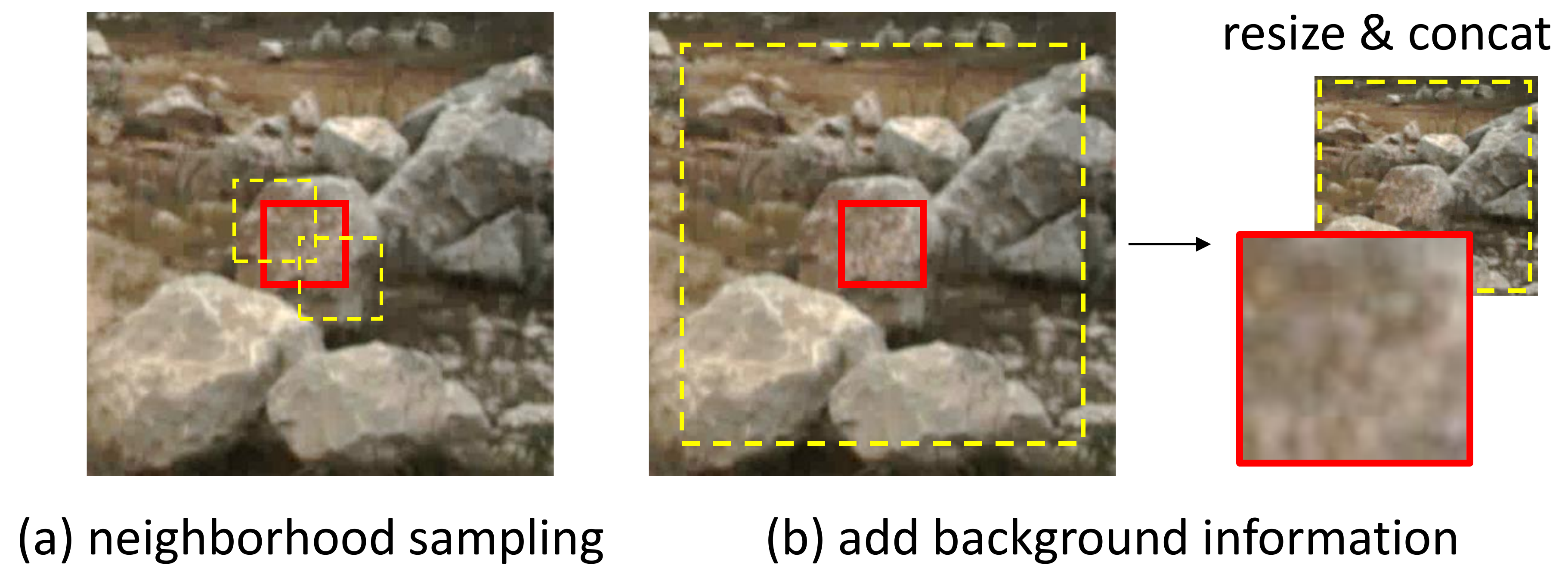}
		\caption{Illustration of (a) neighborhood sampling strategy, and (b) how to add background information with the foreground image patch.}
		\label{fig:dataaug}
	\end{figure}
	
	\subsection{Feature Representation through Contrastive Learning}
	
	\subsubsection {Sampling Strategy}
	
	In each training step, an anchor patch $\mathcal{P}_{k,i}$ is selected to compose a query sample $q$, then a positive sample $q^+$ and $n$ negative samples $\{q^-_i|i=1,..,n\}$ are subsequently composed on the anchor patches of the same image $I_k$.
	
	Based on the label $a_{k,i}$ of $\mathcal{P}_{k,i}$, the anchor patches in the same image $I_k$ are divided into two sets, where $\{\mathcal{P}_{k,i}^+\}$ denotes those sharing the same label $a_{k,i}$, whereas $\{\mathcal{P}_{k,i}^-\}$ for the rest.
	Assume that an off-road scene is spatially continuous, i.e. nearby regions could be semantically similar. An anchor patch $p$ is first randomly selected from $\{\mathcal{P}_{k,i}^+\}$, where an image patch is randomly clipped from $p$'s neighborhood to compose a positive sample $q^+$. As illustrated in Fig.~\ref{fig:dataaug}(a), the randomly clipped neighborhood patches should have the center points within the original one. Similarly, $n$ negative samples $\{q^-_i\}$ are composed on $\{\mathcal{P}_{k,i}^-\}$.
	
	\subsubsection{Composing Sample Data}
	
	As shown in Fig.~\ref{fig:dataaug}(b), sample data contains foreground and background image patches to describe both local and global features. The foreground is image patch $p$, while the background is centered at $p$ but with a larger region to provide global scene context. The foreground and background patches are firstly resized to the same scale, then concatenated along the channel dimension to compose a 6-channel tensor.
	With an image patch $p$, sample data is composed in the same way for the query, positive and negative samples.
	
	In order to improve robustness in diverse scenes, data augmentation (denoted by $\varphi$ in Fig.~\ref{fig:pipeline}) is conducted on the 6-channel tensor of each sample data before forwarding it to the network of $f_{\theta}$. In this research, data augmentation includes random flip, random greyscale, and color jitter, which randomly changes the brightness, contrast, and saturation of an image.
	
	\subsubsection{Network Design and Loss Function}
	A CNN backbone network in practical terms, i.e. AlexNet~\cite{krizhevsky2012imagenet} is used to model $f_{\theta}$, which converts the 6-channel tensor of a query, positive or negative sample to a normalized low-dimensional feature vector $z\in \mathbb{Z}^D$.
	Contrastive learning is used to find $\theta$ in $f_{\theta}$, with which the exponential cosine similarity of the $z$ are high if they share the same labels, whereas low for those differences.
	Following the principle of previous contrastive learning studies~\cite{he2020momentum}, a contrastive loss function InfoNCE~\cite{oord2018representation} is implemented:
	
	\begin{equation}\label{loss}
	L=-\log {\dfrac{\exp (z^T \cdot z^+/\tau)}{\exp (z^T \cdot z^+/\tau)+\sum_{i=1}^{n}{\exp (z^T \cdot z_i^-/\tau)}}}
	\end{equation}
	where $\tau$ denotes a temperature hyper-parameter.
	
	In this work, since the positive and negative samples are comparable only in the same image, the limited quantity makes it possible to get feature representations with reasonable memory consumption. In practice, unlike the typical contrastive learning studies~\cite{Wu_2018_CVPR} using a memory bank to store feature vectors for each training sample, we randomly select positive/negative samples and calculate their features at each training step.
	
	\subsection{Off-road Semantic Segmentation and Mapping}
	
	As illustrated in Fig.~\ref{fig:pipeline}, the workflow contains offline learning and online prediction, while the latter is composed of further two steps: semantic segmentation of single images and semantic mapping using multiple images.
	
	\subsubsection{Offline Learning}
	Given a set of training images $\mathcal{I}=\{I_k\}$ with anchor patches $\mathcal{A}=\{A_k\}$ on each of them, a feature encoder $f_\theta$ is thus learned to convert each image patch to a normalized low-dimensional vector $z\in \mathbb{Z}^D$ in the space of $\mathbb{Z}^D$, the image patches with the same labels are projected close, whereas far for the others.
	
	The $z$ of the anchor patches are then clustered by the K-means method, where a set of mean points $\mathcal{C}=\{\tilde{z}_c\}$ are extracted, representing the features of $\mathcal{K}$ dominant semantic clusters. Here, clustering number $\mathcal{K}$ is a hyper-parameter, which decides the granularity of semantic segmentation.
	
	\subsubsection{Semantic Segmentation}
	
	Given the current image $\mathcal{I}_t$, semantic segmentation $\mathcal{S}_t$ is conducted by generating image patches using sliding windows and predicting a semantic label for each image patch.
	Given an image patch $p_i$, a semantic label is predicted as follows. A 6-channel tensor data is first composed, containing both local and global features of the image patch. The data is then projected by $f_{\theta}$ to a normalized lower-dimensional feature vector $z_i$, which is subsequently compared with the set of feature vectors $\mathcal{C}=\{\tilde{z}_c\}$ representing the $\mathcal{K}$ dominant semantic labels. The image patch is assigned as the semantic label $l_i$ that has the best match on its feature vector, i.e. $l_i = \arg\min_c exp(z_i^T \cdot \tilde{z}_c)$.
	
	To make up denser semantic segmentation, we could adjust the step size of sliding windows. For example, we can assign the semantic label to $3*3$ pixels centered at each image patch, while setting sliding windows' horizontal/vertical step size to 3 pixels, then get denser semantic segmentation results.
	
	\subsubsection{Semantic Mapping} \label{3_SM}
	
	Centered at the ego vehicle's location at the frame, a horizontal plane is drawn at the ground level and tessellated into regular grids.  
	The pixel labels of the current image can be projected onto the grids with the camera’s calibration parameters. 
	Besides, the pixel labels of early frames can also be projected onto the grids with additionally the vehicle's localization data at each frame.
	If a 3D LiDAR is associated, the projection can be conducted via LiDAR points, where the up and down of off-road terrain can be taken into calculation. 
	Since a single grid can have multiple label predictions, let $\sigma_{x,y}^c$ denote the counts of predicting label $c$ of grid $(x,y)$, the semantic label $l_{x,y}=\mathop{\text{argmax}_{c} (\sigma_{x,y}^c)}$ is assigned to the grid. Meanwhile, a confidence map is estimated to indicate the confidence of predicted labels. The confidence value of grid $(x,y)$ is assigned as $\max(\sigma_{x,y}^c)/\sum{\sigma_{x,y}^c}$, which can also serve as a measure to evaluate prediction consistency.
	
	\section{Experimental Results}	\label{exp}
	\subsection{Experimental Data}
	An off-road dataset is developed to evaluate the proposed method.
	The dataset is collected by an instrumented vehicle with a front-view monocular RGB camera,
	a GPS/IMU suite, and a 3D LiDAR. In this work, we use visual images for semantic segmentation, 
	while GPS/IMU provides 6 DoF poses of the ego vehicle, which is used for mapping. 
	3D LiDAR is mainly used to examine the semantic meaning of the fine-grained segmentation, 
	while it is also used in this research in projecting visual labels to a horizontal plane so as to generate a more accurate map by considering the up and down of off-road terrain.
	
	\begin{table}[]
		\centering
		\caption{Statistics of the off-road dataset}
		\label{tab:dataset}
		\begin{tabular}{cccc} 
			\hline
			& subset A & subset B & subset C  \\ 
			\hline
			total frames        & 5064     & 3239     & 4098      \\
			frames for training & 50       & 100      & 80        \\
			anchors       & 973      & 1606     & 1437      \\
			\hline
		\end{tabular}
	\end{table}
	
	\begin{figure}[]
		\centering
		\includegraphics[scale=0.28]{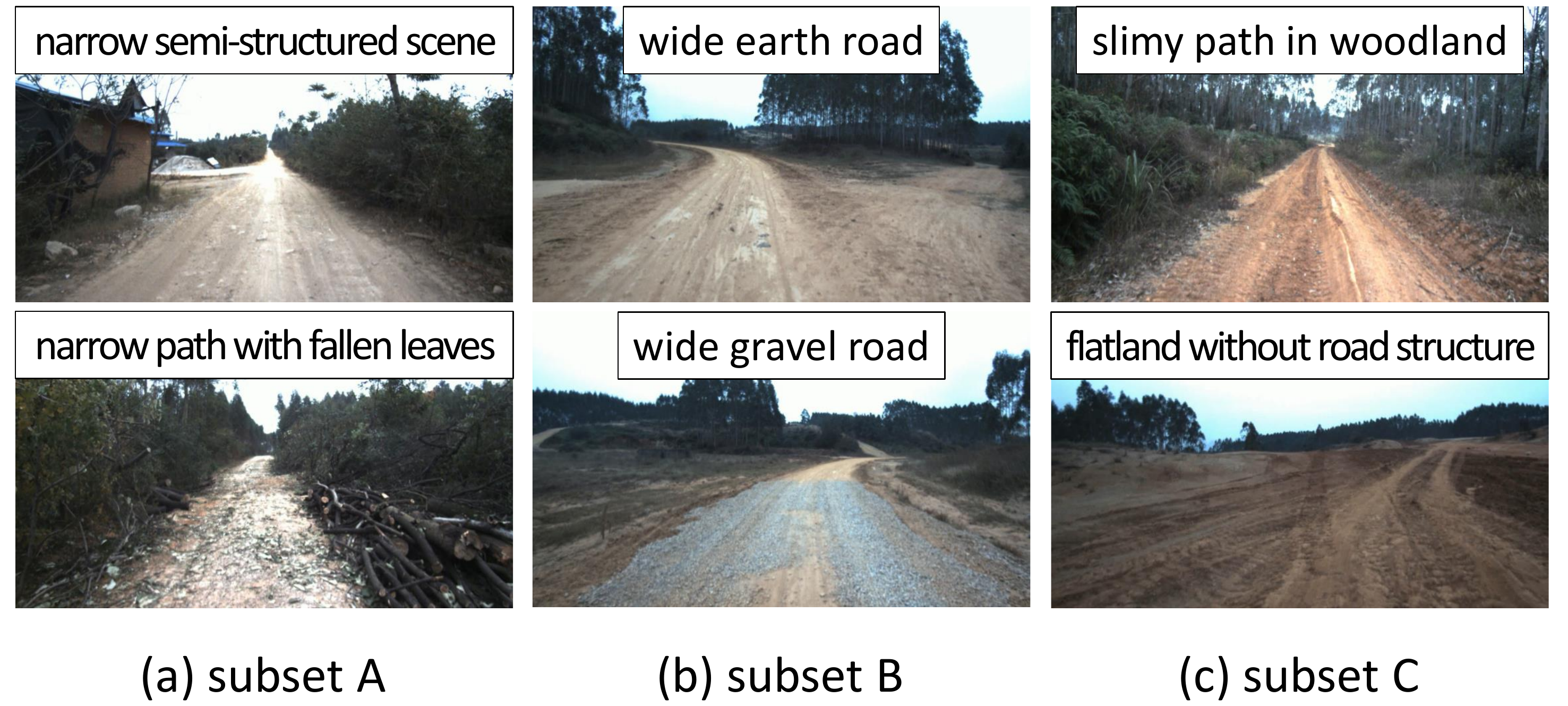}
		\caption{Typical scenes in three sub-datasets, which include diverse off-road scenarios.}
		\label{fig:dataset}
	\end{figure}
	
	As shown in Table~\ref{tab:dataset}, the dataset contains over 12000 image frames of three driving segments that represent very diverse off-road scenes.
	Take subset A as an example, 50 image frames are randomly selected, which account for 10\% of the total 5064 frames of subset A. 
	973 anchor patches are annotated on the 50 image frames by a human operator, which are used in training.
	The rest image frames of subset A are used in testing, 
	and the image frames of subset B and C are also used to test the model trained on subset A in the experiment of cross-scene validation.
	Experiments on subset B and C are conducted in the same way to examine the results of semantic segmentation.
	To this end, image frames are used for testing and image patches are manually annotated in the same way as the anchors, which are used as \textit{ground truth} to evaluate the accuracy of the results.
	
	The three subsets contain driving data at very different off-road scenes. 
	As illustrated in Fig.~\ref{fig:dataset}, the scenarios in
	subset A are mostly narrow roads with bushes aside, subset B are relatively wide scenes, and subset C includes diverse
	scenarios like slimy paths in woodland and flatland without
	road structure. In the experiments, we train and test the
	proposed method on different subsets to evaluate its cross-scene generalization performance.
	
	\subsection{Evaluation Metrics}
	
	\begin{figure*}[]
		\centering
		\includegraphics[width=0.9\textwidth]{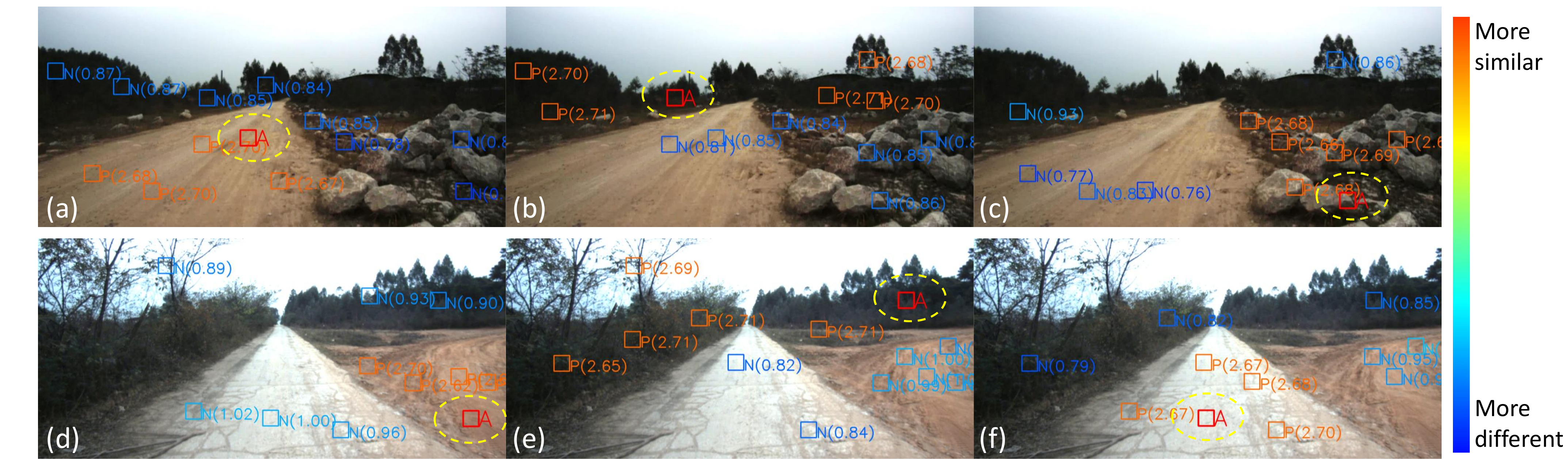}
		\caption{Visualization of feature similarity of query anchor (A) to its positive (P) and negative (N) samples. Query anchors are circled by yellow rings. Numbers in parentheses measure samples' exponential cosine similarity to the query anchor.}
		\label{fig:anchor_dis}
	\end{figure*}
	
	Suppose that there are $N$ anchors in one frame, then any two anchors must be either positive or negative samples of each other. Hence, there exists $N \cdot(N-1)$ pairs anchor constraints.
	We denote positive samples' constraints as $Pos(i,j)$. If anchor patch $\mathcal{P}_{k,i}$ and $\mathcal{P}_{k,j}$ are positive samples of each other and classified into the same semantic cluster, then $Pos(i,j)=1$. Otherwise, if they are not classified to the same semantic cluster, $Pos(i,j)=0$.
	Negative samples' constraints are defined in a similar way and denoted as $Neg(i,j)$.
	
	We use the following metrics called \textit{anchor accuracy} to evaluate how well the clustering results fit human annotations:
	\begin{equation}
	\label{R}
	\mathcal{R}=\dfrac{\sum_{i,j}{Pos(i,j)}+\sum_{i,j}{Neg(i,j)}}{N\cdot (N-1)}, ~i \neq j
	\end{equation}
	Essentially, it can be seen as Rand Index~\cite{rand1971objective}, which is a commonly used measurement for clustering.
	
	\begin{table*}
		\centering
		\caption{Cross-Scene Validation Results ($\mathcal{R}$) on Different Datasets}
		\label{tab:cross_eval}
		\renewcommand{\arraystretch}{1.2}
		\begin{threeparttable}
			\begin{tabular}{ccc|ccc|ccc|ccc!{\color{black}\vrule}c} 
				\hline
				\multirow{3}{*}{model} & \multirow{3}{*}{\makecell[c]{data\\ aug.}} & \multirow{3}{*}{\makecell[c]{BG\\ size}} & \multicolumn{3}{c|}{train on subset A}                                                                             & \multicolumn{3}{c|}{train on subset B}                                                                             & \multicolumn{3}{c|}{train on subset C}                                                                             & \multirow{3}{*}{\makecell[c]{$\bar{\mathcal{R}}$ on \\ test sets}}                      \\ 
				\cline{4-12}
				&                            &                             & \multicolumn{3}{c|}{test on}                                                                                       & \multicolumn{3}{c|}{test on}                                                                                       & \multicolumn{3}{c|}{test on}                                                                                       &                                                      \\
				&                            &                             & A      & B                                                   & C                                                   & B      & A                                                   & C                                                   & C      & A                                                   & \multicolumn{1}{c|}{B}                              &                                                      \\ 
				\hline
				base                   & $\times$                          & $\times$                           & 0.9854 & {\cellcolor[rgb]{1,0.906,0.906}}0.8548              & {\cellcolor[rgb]{0.992,0.722,0.722}}0.8509          & 0.9997 & {\cellcolor[rgb]{1,0.906,0.906}}0.7957              & {\cellcolor[rgb]{1,0.906,0.906}}0.8492              & 0.9966 & {\cellcolor[rgb]{1,0.906,0.906}}0.8288              & {\cellcolor[rgb]{0.992,0.745,0.749}}0.9258          & {\cellcolor[rgb]{1,0.906,0.906}}0.8509               \\
				base\_DA               & \checkmark                          & $\times$                          & 0.9693 & {\cellcolor[rgb]{0.996,0.773,0.773}}0.8792          & {\cellcolor[rgb]{1,0.906,0.906}}0.8422              & 0.9959 & {\cellcolor[rgb]{0.992,0.733,0.733}}0.8210          & {\cellcolor[rgb]{0.992,0.745,0.749}}0.8625          & 0.9913 & {\cellcolor[rgb]{1,0.898,0.898}}0.8296              & {\cellcolor[rgb]{1,0.906,0.906}}0.9119              & {\cellcolor[rgb]{0.996,0.835,0.835}}0.8578           \\
				BG192                  & \checkmark                          & 192                         & 0.9939 & {\cellcolor[rgb]{0.976,0.471,0.478}}0.9330          & {\cellcolor[rgb]{0.973,0.412,0.42}}\textbf{0.8650}  & 0.9994 & {\cellcolor[rgb]{0.98,0.514,0.518}}0.8524           & {\cellcolor[rgb]{0.973,0.412,0.42}}\textbf{0.8899}  & 0.9944 & {\cellcolor[rgb]{0.98,0.537,0.545}}0.8653           & {\cellcolor[rgb]{0.98,0.502,0.51}}0.9468            & {\cellcolor[rgb]{0.976,0.475,0.482}}0.8920           \\
				BG256                  & \checkmark                          & 256                         & 0.9987 & {\cellcolor[rgb]{0.976,0.455,0.463}}0.9360          & {\cellcolor[rgb]{0.976,0.463,0.471}}0.8627          & 0.9991 & {\cellcolor[rgb]{0.976,0.475,0.482}}0.8577          & {\cellcolor[rgb]{0.98,0.486,0.494}}0.8839           & 0.9934 & {\cellcolor[rgb]{0.98,0.525,0.533}}0.8665           & {\cellcolor[rgb]{0.976,0.451,0.459}}0.9512          & {\cellcolor[rgb]{0.976,0.467,0.471}}0.8930           \\
				BG320                  & \checkmark                          & 320                         & 0.9986 & {\cellcolor[rgb]{0.973,0.412,0.42}}\textbf{0.9433}  & {\cellcolor[rgb]{0.984,0.612,0.616}}0.8559          & 0.9980 & {\cellcolor[rgb]{0.973,0.412,0.42}}\textbf{0.8667}  & {\cellcolor[rgb]{0.976,0.42,0.427}}0.8895           & 0.9958 & {\cellcolor[rgb]{0.973,0.412,0.42}}\textbf{0.8776}  & {\cellcolor[rgb]{0.973,0.412,0.42}}\textbf{0.9544}  & {\cellcolor[rgb]{0.973,0.412,0.42}}\textbf{0.8979}   \\
				\hline
			\end{tabular}
			\begin{tablenotes}
				\footnotesize
				\item[*] \textbf{BG}: background; \textbf{base}: pipeline without data augmentation or background information; \textbf{base\_DA}: use data augmentation, without background information; \textbf{BG192/256/320}: complete pipeline with different background size; \textbf{$\bar{\mathcal{R}}$}: average anchor accuracy $\mathcal{R}$.
			\end{tablenotes}
		\end{threeparttable}
		\vspace{-3mm}
	\end{table*}
	
	\subsection{Results on Proposed Method}
	To evaluate the proposed method, we design the following experiments: (1) feature similarity measurement, explore the validity of feature encoder and similarity measurement learned by contrastive learning. (2) cross-scene validation and ablation study, verify the performance and robustness of our proposed method in diverse test scenes while exploring the effects of different module settings. (3) fine-grained semantic segmentation and mapping, make concrete case study and statistical analysis from additional LiDAR data to show the validity of our fine-grained results.

	\begin{figure}[]
		\centering
		\includegraphics[scale=0.25]{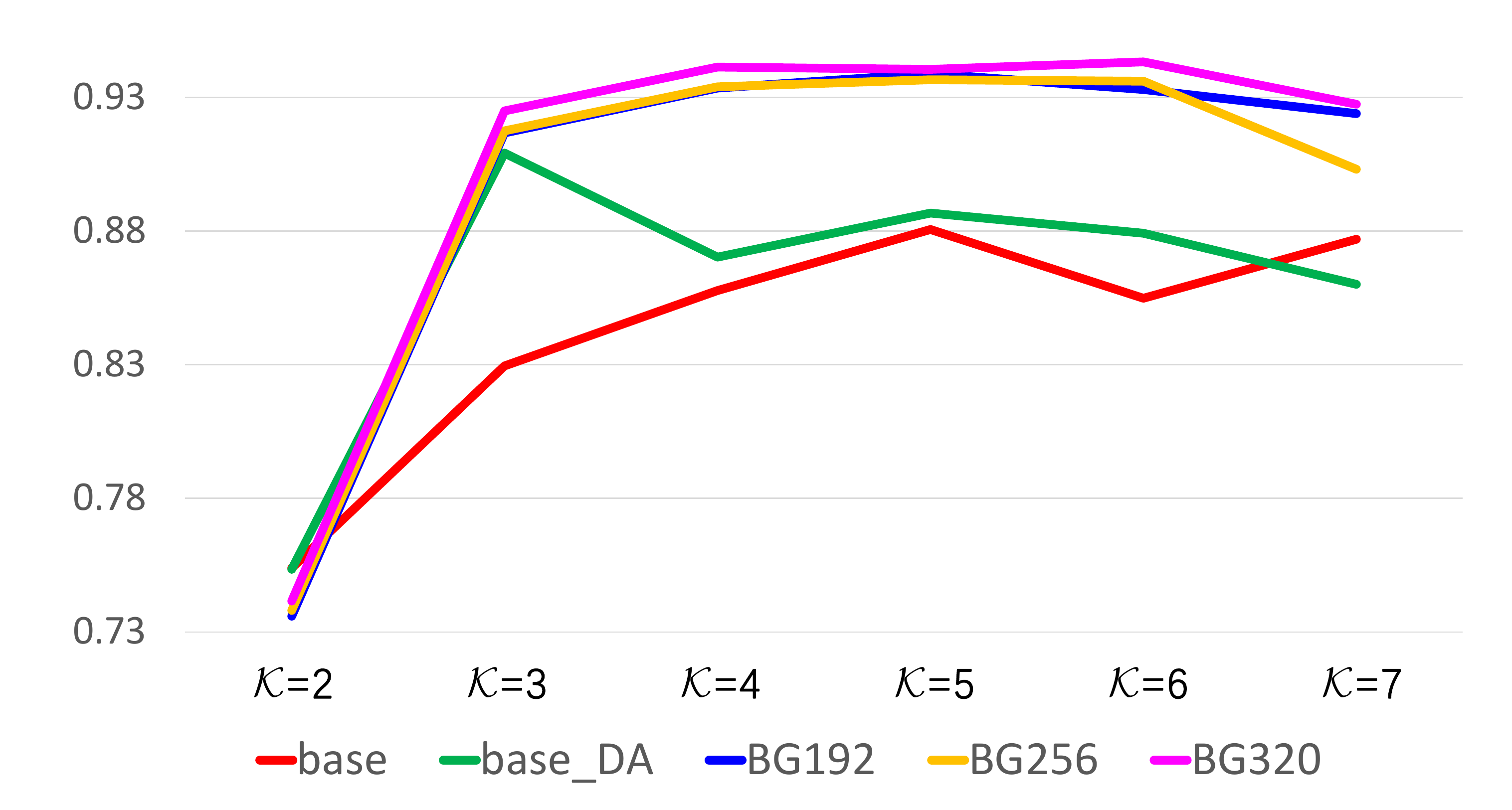}
		\caption{Average $\mathcal{R}$ of models under different clustering number $\mathcal{K}$.}
		\label{fig:kmeans_exp}
	\end{figure}
	
	\subsubsection{Feature Similarity Measurement}
	The feature encoder $f_\theta$ aims to make similar image patches closer and different image patches farther in feature space.
	In Fig.~\ref{fig:anchor_dis}, we visualize some concrete cases of the learned similarity measurement $sim(p_i,p_j)=exp(z_i^T \cdot z_j)$. In all images, the query anchors are circled by yellow rings, while the other anchor patches are randomly sampled and colorized by their exponential cosine similarity to the query anchor. For example, in Fig.~\ref{fig:dataaug}(a), the query anchor is located on the earth road. We can find that patches on the earth road are closer to red, and other patches located on different semantic areas are generally blue, which indicates the lower similarity to the query anchor. The feature similarity distribution is in accord with the semantic differences. Similar situations are general in other images. \textbf{As a result, the feature encoder and similarity measurement learned by contrastive learning are able to distinguish similar or different image patches.}
	
	\subsubsection{Cross-Scene Validation and Ablation Study}
	
	%
	%
	For comprehensive evaluations of the proposed method, we make cross-scene validation on models with different settings, and the statistics are shown in Table~\ref{tab:cross_eval}. The table cells are colorized along column data when training and testing on different subsets. The last column lists the average anchor accuracy $\bar{\mathcal{R}}$ on test sets (the subsets different with the training one). It is obvious that \textit{BG320} has the best performance on test sets, and all three models with background information have $\mathcal{R}$ over 85\% among all conditions, which demonstrates the robustness of our proposed method.
	The data augmentation and background information can both increase models' performance, while the latter makes more contribution. Increasing background size could sightly improve overall performance, but is not obvious in all situations.
	
	\begin{figure*}[]
		\centering
		\includegraphics[width=0.88\textwidth]{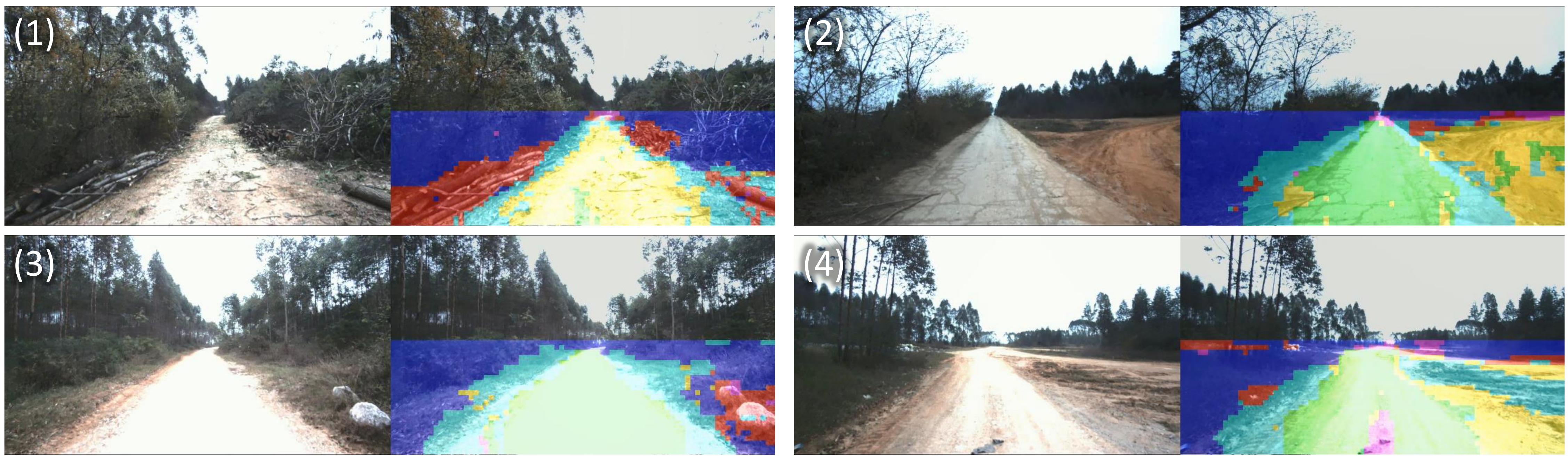}
		\caption{Some results of fine-grained semantic segmentation.}
		\label{fig:semantic_segmentation}
	\end{figure*}
	
	\begin{figure*}[]
		\centering
		\includegraphics[width=0.95\textwidth]{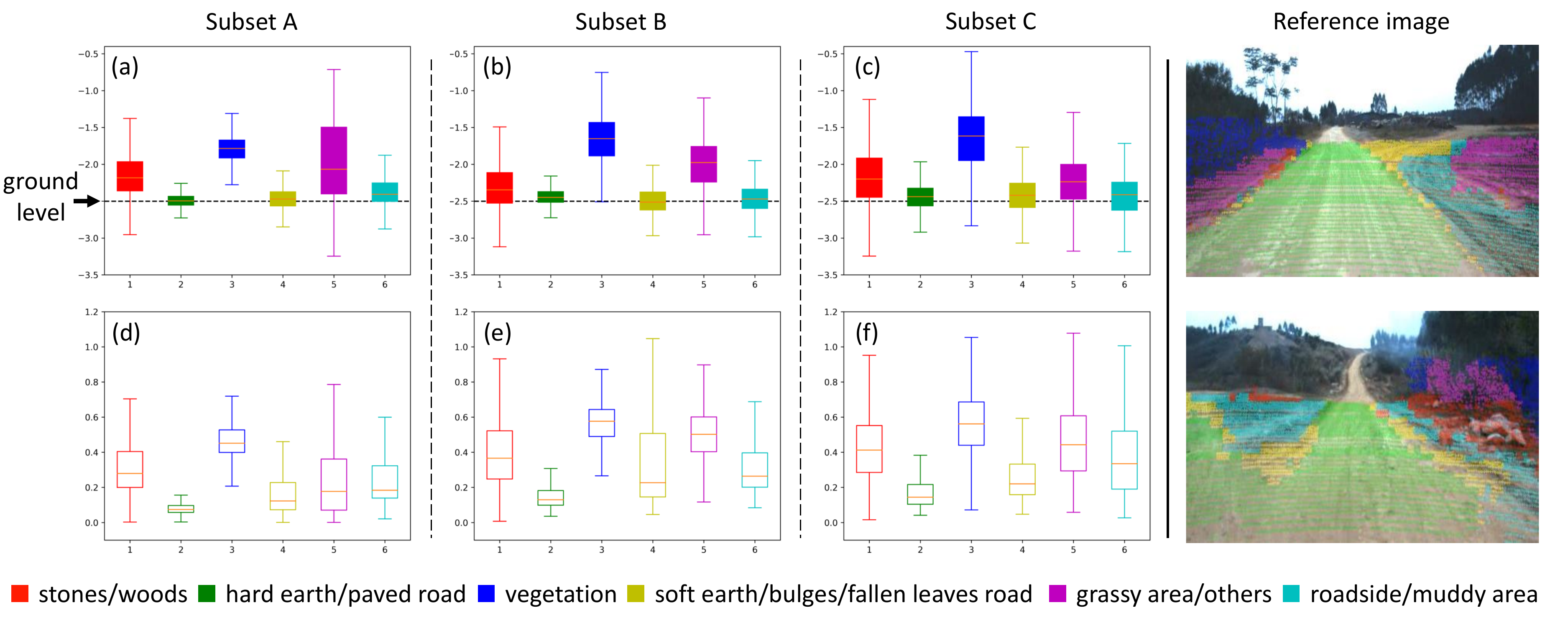}
		\caption{Traversability analysis of semantic clusters by 3D LiDAR data. (a-c) boxplots of points average height, indicate height distribution of different categories. (d-f) boxplots of points height variance, indicate surface flatness and traversability cost. }
		\label{fig:lidar_analysis}
		\vspace{-5mm}
	\end{figure*}

	\begin{figure*}[]
		\centering
		\includegraphics[width=0.85\textwidth]{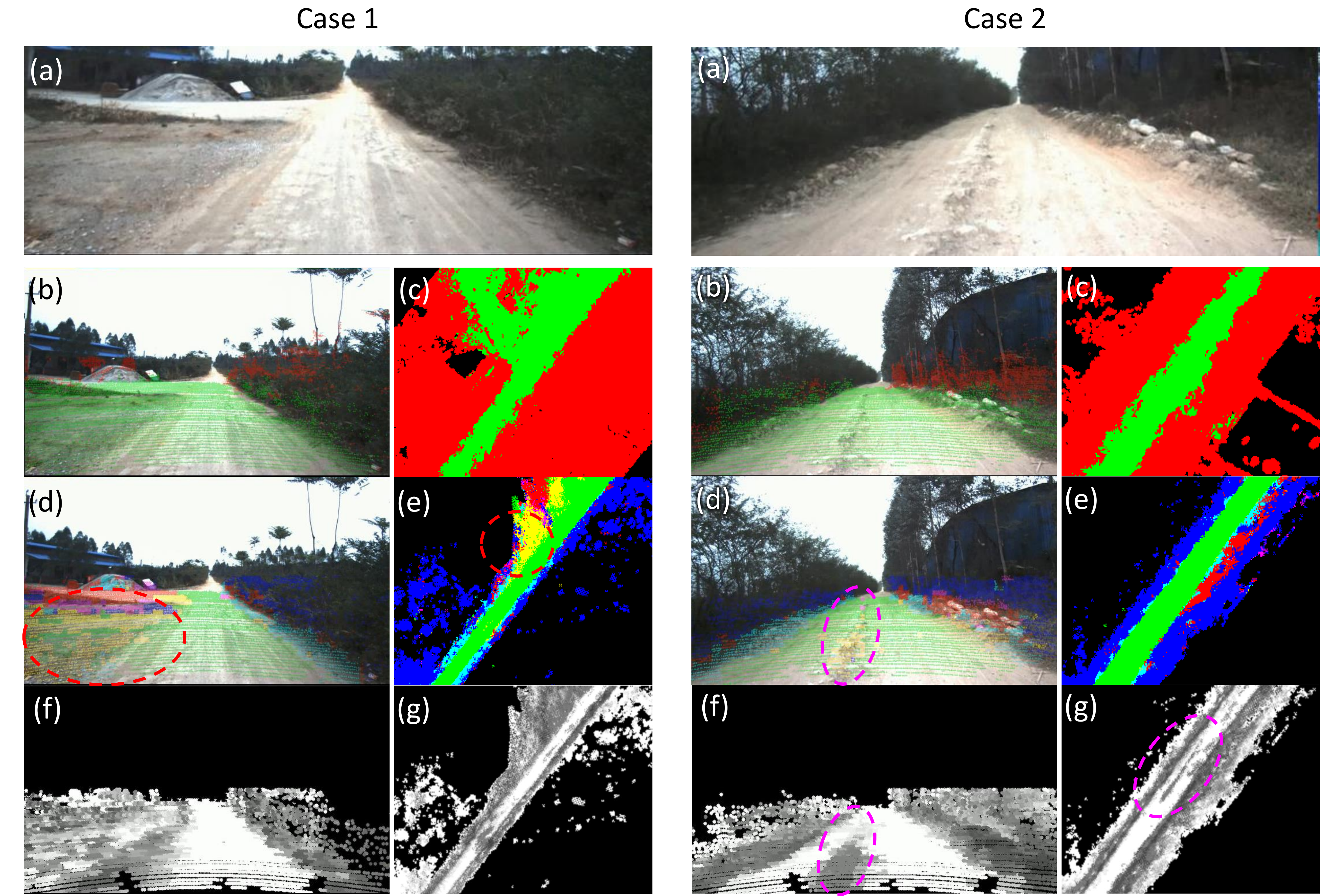}
		\caption{Case study of fine-grained semantic map and confidence map, compared with coarse-grained  results. (a) scene image. (b) coarse-grained segmentation (binary classification). (c) coarse-grained semantic map (bird's-eye-view). (d) fine-grained semantic segmentation (projected on point clouds). (e) fine-grained semantic map (bird's-eye-view). (f) confidence map projected to camera-view, whiter pixels indicate higher confidence. (g) bird's-eye-view confidence map.}
		\label{fig:semantic_mapping}
		\vspace{-4mm}
	\end{figure*}
	
	To explore how clustering number $\mathcal{K}$ affects models' performance, an ablation study is made as shown in Fig.~\ref{fig:kmeans_exp}. We can find that the models' performance with regard to $\mathcal{K}$ are basically stable when $\mathcal{K} \geq 4$, and slightly decrease when $\mathcal{K}>6$. In general, models' performance approximately orders the same as Table~\ref{tab:cross_eval}. Therefore, we choose $\mathcal{K}=6$ as other experiments' setting to balance the fine-grained demand and model performance.
	
	\textbf{In summary, the proposed method achieves 89.8\% average anchor accuracy in cross-scene validation, and the performance is stable with regard to different clustering numbers, which demonstrates the robustness and generalization of our method.}

	\subsubsection{Fine-Grained Semantic Segmentation and Mapping}
	
	Due to the absence of pixel-level annotations for the task, we next demonstrate the validity of our fine-grained results through case studies and additional LiDAR data analysis. The following results are all based on the model trained by 50 frames of subset A.
	
	Fig.~\ref{fig:semantic_segmentation} shows some cases of fine-grained semantic segmentation. Because this work focuses on off-road traversability analysis, so only the bottom half of the image is predicted for simplicity. The semantic labels are not pre-designed, but we can find their intrinsic meanings through these concrete cases. For example, \textit{green} indicates hard earth road and paved road, \textit{blue} pixels are vegetation, \textit{yellow} pixels are road with fallen leaves or soft earth, \textit{red} pixels are stones or woods, etc. Different clusters can generally distinguish diverse semantic meanings.
	
	Statistical analysis is provided in Fig.~\ref{fig:lidar_analysis}, which is based on 3D LiDAR data with labels projected from image semantic segmentation.
	In Fig.~\ref{fig:lidar_analysis}(a-c), three categories' terrain elevation (\textit{green}, \textit{yellow}, and \textit{cyan}) mainly distribute around the ground level, which are three primary road types. Furthermore, from Fig.~\ref{fig:lidar_analysis}(d-f), we can find their different traversability cost. The \textit{green} boxes have the narrowest variance distribution, corresponding to the most easily passable paved road and hard earth. The \textit{yellow} and \textit{cyan} boxes are longer, indicating bumpier road surface. The \textit{blue} boxes indicate bushes and trees, with the highest elevation and traversability cost.
	\textbf{In a word, examined by associated 3D LiDAR data, the fine-grained segments of images are proved to have different levels of toughness and terrain elevation, which represents their semantical meaningfulness.}
	
	The semantic maps and confidence maps are shown in Fig.~\ref{fig:semantic_mapping}. In case 1, our fine-grained predictions label the roadside area (\textit{yellow}) with higher traversability cost than middle road (\textit{green}). In case 2, bulges in the middle of the road are separated in the single frame segmentation, but not stable enough to obtain majority votes in the semantic map. The confidence map can be helpful to distinguish this subtle traversability difference, where the bumpy area is darker than other flat roads. \textbf{Therefore, the resultant fine-grained semantic maps and confidence maps can provide rich information for robots to traverse in complex off-road scenes.}
	
	\begin{figure*}[]
		\centering
		\includegraphics[width=0.85\textwidth]{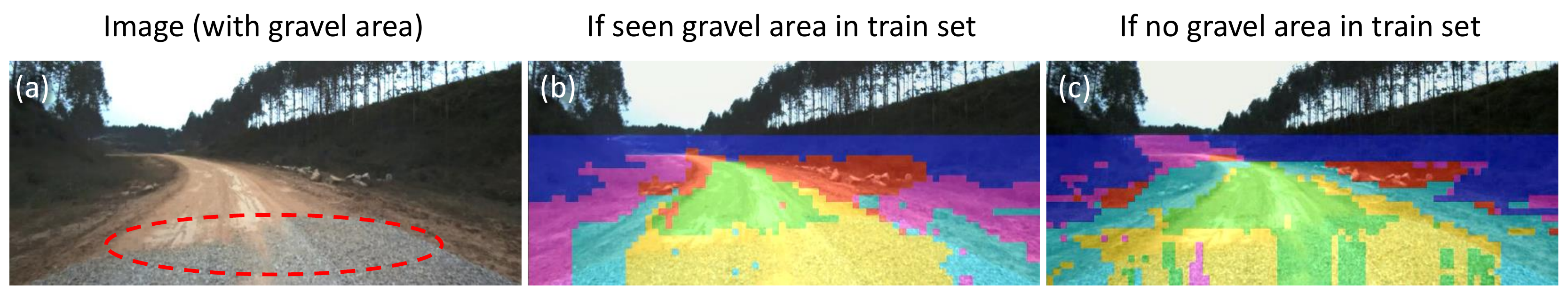}
		\caption{A challenging case: when meeting unseen semantic categories.}
		\label{fig:challenges}
		\vspace{-4mm}
	\end{figure*}
	
	\subsection{Challenges}
	Currently, there are still some challenges with the proposed method. Firstly, the current pipeline to obtain dense predictions has a high computational cost. Predicting one image patch takes about 30 ms on an NVIDIA TITAN X. Although parallel computing helps to predict more patches at one time, the repetitive computation of overlapped patches can be optimized by temporal and spatial consistency in future works.
	The second one is unseen semantic categories, or called out of distribution (OOD) samples, as shown in Fig.~\ref{fig:challenges}. The current pipeline will not discriminate unseen category samples, but simply classified them into existing semantic clusters, which may lead to confused predictions as Fig.~\ref{fig:challenges}(c). To minimize labor costs, the OOD sample detection and incremental training mechanism deserve to be explored in our future works.

	\section{CONCLUSIONS}	\label{conclusions}
	In this paper, we propose a contrastive learning based method for off-road fine-grained semantic segmentation and mapping. With a set of human-annotated anchor patches, a feature representation is learned to discriminate regions with different traversability. After that, the fine-grained semantic segmentation and mapping pipeline is proposed for off-road scene understanding.
	For the experimental study of our method, we develop an off-road dataset with three driving segments that represent very diverse off-road scenes. The proposed method achieves 89.8\% anchor accuracy in cross-scene validation by evaluating the matching with human-annotated image patches.
	Examined by associated 3D LiDAR data, the fine-grained segments of visual images are demonstrated to have different levels of toughness and terrain elevation, which represents their semantical meaningfulness.
	The resultant maps contain both fine-grained labels and confidence values, providing rich information to support a robot traversing complex off-road scenes.
	Future work will be addressed on improving the computational efficiency by temporal and spatial consistency, and exploring OOD sample detection mechanism and incremental learning ability for long-term deployment on off-road robots.
	
	


	
	
	

	
	
	\printbibliography

@inproceedings{behley2019semantickitti,
	title={{SemanticKITTI}: A dataset for semantic scene understanding of {LiDAR} sequences},
	author={Behley, Jens and Garbade, Martin and Milioto, Andres and Quenzel, Jan and Behnke, Sven and Stachniss, Cyrill and Gall, Jurgen},
	booktitle={International Conference on Computer Vision},
	pages={9297--9307},
	year={2019},
	organization={IEEE}
}

@article{rateke2019passive,
	title={Passive vision region-based road detection: A literature review},
	author={Rateke, Thiago and Justen, Karla A and Chiarella, Vito F and Sobieranski, Antonio C and Comunello, Eros and Wangenheim, Aldo Von},
	journal={ACM Computing Surveys},
	volume={52},
	number={2},
	pages={1--34},
	year={2019},
	publisher={ACM New York, NY, USA}
}

@article{zhou2012self,
	title={Self-supervised learning to visually detect terrain surfaces for autonomous robots operating in forested terrain},
	author={Zhou, Shengyan and Xi, Junqiang and McDaniel, Matthew W and Nishihata, Takayuki and Salesses, Phil and Iagnemma, Karl},
	journal={Journal of Field Robotics},
	volume={29},
	number={2},
	pages={277--297},
	year={2012},
	publisher={Wiley Online Library}
}

@article{ososinski2015automatic,
	title={Automatic Driving on Ill-defined Roads: An Adaptive, Shape-constrained, Color-based Method},
	author={Ososinski, Marek and Labrosse, Fr{\'e}d{\'e}ric},
	journal={Journal of Field Robotics},
	volume={32},
	number={4},
	pages={504--533},
	year={2015},
	publisher={Wiley Online Library}
}

@inproceedings{nefian2006detection,
	title={Detection of drivable corridors for off-road autonomous navigation},
	author={Nefian, Ara V and Bradski, Gary R},
	booktitle={International Conference on Image Processing},
	pages={3025--3028},
	year={2006},
	organization={IEEE}
}

@inproceedings{siam2017deep,
	title={Deep semantic segmentation for automated driving: Taxonomy, roadmap and challenges},
	author={Siam, Mennatullah and Elkerdawy, Sara and Jagersand, Martin and Yogamani, Senthil},
	booktitle={International Conference on Intelligent Transportation Systems},
	pages={1--8},
	year={2017},
	organization={IEEE}
}

@article{oord2018CPC,
	title={Representation learning with contrastive predictive coding},
	author={Oord, Aaron van den and Li, Yazhe and Vinyals, Oriol},
	journal={arXiv preprint arXiv:1807.03748},
	year={2018}
}

@inproceedings{chen2020simple,
	title={A simple framework for contrastive learning of visual representations},
	author={Chen, Ting and Kornblith, Simon and Norouzi, Mohammad and Hinton, Geoffrey},
	booktitle={International Conference on Machine Learning},
	pages={1597--1607},
	year={2020},
	organization={PMLR}
}

@inproceedings{he2020momentum,
	title={Momentum contrast for unsupervised visual representation learning},
	author={He, Kaiming and Fan, Haoqi and Wu, Yuxin and Xie, Saining and Girshick, Ross},
	booktitle={Conference on Computer Vision and Pattern Recognition},
	pages={9729--9738},
	year={2020}
}

@article{badue2020self,
	title={Self-driving cars: A survey},
	author={Badue, Claudine and Guidolini, R{\^a}nik and Carneiro, Raphael Vivacqua and Azevedo, Pedro and Cardoso, Vinicius Brito and Forechi, Avelino and Jesus, Luan and Berriel, Rodrigo and Paixao, Thiago Meireles and Mutz, Filipe and others},
	journal={Expert Systems with Applications},
	pages={113816},
	year={2020},
	publisher={Elsevier}
}

@article{feng2020deep,
	title={Deep multi-modal object detection and semantic segmentation for autonomous driving: Datasets, methods, and challenges},
	author={Feng, Di and Haase-Schuetz, Christian and Rosenbaum, Lars and Hertlein, Heinz and Glaeser, Claudius and Timm, Fabian and Wiesbeck, Werner and Dietmayer, Klaus},
	journal={Transactions on Intelligent Transportation Systems},
	year={2020},
	publisher={IEEE}
}

@article{krizhevsky2012imagenet,
  title={{ImageNet} classification with deep convolutional neural networks},
  author={Krizhevsky, Alex and Sutskever, Ilya and Hinton, Geoffrey E},
  journal={Advances in Neural Information Processing Systems},
  volume={25},
  pages={1097--1105},
  year={2012}
}

@InProceedings{Wu_2018_CVPR,
	author = {Wu, Zhirong and Xiong, Yuanjun and Yu, Stella X. and Lin, Dahua},
	title = {Unsupervised Feature Learning via Non-Parametric Instance Discrimination},
	booktitle = {Conference on Computer Vision and Pattern Recognition},
	month = {June},
	year = {2018}
}

@article{wellhausen2019should,
	title={Where should {I} walk? predicting terrain properties from images via self-supervised learning},
	author={Wellhausen, Lorenz and Dosovitskiy, Alexey and Ranftl, Ren{\'e} and Walas, Krzysztof and Cadena, Cesar and Hutter, Marco},
	journal={Robotics and Automation Letters},
	volume={4},
	number={2},
	pages={1509--1516},
	year={2019},
	publisher={IEEE}
}

@article{oord2018representation,
	title={Representation learning with contrastive predictive coding},
	author={Oord, Aaron van den and Li, Yazhe and Vinyals, Oriol},
	journal={arXiv preprint arXiv:1807.03748},
	year={2018}
}

@article{rand1971objective,
	title={Objective criteria for the evaluation of clustering methods},
	author={Rand, William M},
	journal={Journal of the American Statistical Association},
	volume={66},
	number={336},
	pages={846--850},
	year={1971},
	publisher={Taylor \& Francis Group}
}

@inproceedings{kong2009vanishing,
	title={Vanishing point detection for road detection},
	author={Kong, Hui and Audibert, Jean-Yves and Ponce, Jean},
	booktitle={Conference on Computer Vision and Pattern Recognition},
	pages={96--103},
	year={2009},
	organization={IEEE}
}

@article{shi2015fast,
	title={Fast and robust vanishing point detection for unstructured road following},
	author={Shi, Jinjin and Wang, Jinxiang and Fu, Fangfa},
	journal={Transactions on Intelligent Transportation Systems},
	volume={17},
	number={4},
	pages={970--979},
	year={2015},
	publisher={IEEE}
}

@inproceedings{zhou2010self,
	title={Self-supervised learning method for unstructured road detection using fuzzy support vector machines},
	author={Zhou, Shengyan and Iagnemma, Karl},
	booktitle={International Conference on Intelligent Robots and Systems},
	pages={1183--1189},
	year={2010},
	organization={IEEE}
}

@article{jeong2002vision,
	title={Vision-based adaptive and recursive tracking of unpaved roads},
	author={Jeong, Hong and Oh, Yuns and Park, Jeong-Ho and Koo, BS and Lee, Sang Wook},
	journal={Pattern Recognition Letters},
	volume={23},
	number={1-3},
	pages={73--82},
	year={2002},
	publisher={Elsevier}
}

@inproceedings{lu2014hierarchical,
	title={A hierarchical approach for road detection},
	author={Lu, Keyu and Li, Jian and An, Xiangjing and He, Hangen},
	booktitle={International Conference on Robotics and Automation},
	pages={517--522},
	year={2014},
	organization={IEEE}
}

@inproceedings{alon2006off,
	title={Off-road path following using region classification and geometric projection constraints},
	author={Alon, Yaniv and Ferencz, Andras and Shashua, Amnon},
	booktitle={Conference on Computer Vision and Pattern Recognition},
	volume={1},
	pages={689--696},
	year={2006},
	organization={IEEE}
}

@inproceedings{wang2009unstructured,
	title={Unstructured road detection using hybrid features},
	author={Wang, Jian and Ji, Zhong and Su, Yu-Ting},
	booktitle={International Conference on Machine Learning and Cybernetics},
	volume={1},
	pages={482--486},
	year={2009},
	organization={IEEE}
}

@article{mei2017scene,
	title={Scene-adaptive off-road detection using a monocular camera},
	author={Mei, Jilin and Yu, Yufeng and Zhao, Huijing and Zha, Hongbin},
	journal={Transactions on Intelligent Transportation Systems},
	volume={19},
	number={1},
	pages={242--253},
	year={2017},
	publisher={IEEE}
}

@inproceedings{cordts2016cityscapes,
	title={The {CityScapes} dataset for semantic urban scene understanding},
	author={Cordts, Marius and Omran, Mohamed and Ramos, Sebastian and Rehfeld, Timo and Enzweiler, Markus and Benenson, Rodrigo and Franke, Uwe and Roth, Stefan and Schiele, Bernt},
	booktitle={Conference on Computer Vision and Pattern Recognition},
	pages={3213--3223},
	year={2016}
}

@inproceedings{long2015fully,
	title={Fully convolutional networks for semantic segmentation},
	author={Long, Jonathan and Shelhamer, Evan and Darrell, Trevor},
	booktitle={Conference on Computer Vision and Pattern Recognition},
	pages={3431--3440},
	year={2015}
}

@inproceedings{suger2015traversability,
	title={Traversability analysis for mobile robots in outdoor environments: A semi-supervised learning approach based on {3D-LiDAR} data},
	author={Suger, Benjamin and Steder, Bastian and Burgard, Wolfram},
	booktitle={International Conference on Robotics and Automation},
	pages={3941--3946},
	year={2015},
	organization={IEEE}
}

@inproceedings{gao2019off,
	title={Off-Road Drivable Area Extraction Using {3D LiDAR} Data},
	author={Gao, Biao and Xu, Anran and Pan, Yancheng and Zhao, Xijun and Yao, Wen and Zhao, Huijing},
	booktitle={Intelligent Vehicles Symposium},
	pages={1505--1511},
	year={2019},
	organization={IEEE}
}

@article{zurn2020self,
	title={Self-supervised visual terrain classification from unsupervised acoustic feature learning},
	author={Z{\"u}rn, Jannik and Burgard, Wolfram and Valada, Abhinav},
	journal={Transactions on Robotics},
	year={2020},
	publisher={IEEE}
}

@article{sharma2019semantic,
	title={Semantic segmentation with transfer learning for off-road autonomous driving},
	author={Sharma, Suvash and Ball, John E and Tang, Bo and Carruth, Daniel W and Doude, Matthew and Islam, Muhammad Aminul},
	journal={Sensors},
	volume={19},
	number={11},
	pages={2577},
	year={2019},
	publisher={Multidisciplinary Digital Publishing Institute}
}

@inproceedings{tang2017one,
	title={From one to many: Unsupervised traversable area segmentation in off-road environment},
	author={Tang, Li and Ding, Xiaqing and Yin, Huan and Wang, Yue and Xiong, Rong},
	booktitle={International Conference on Robotics and Biomimetics},
	pages={787--792},
	year={2017},
	organization={IEEE}
}

@inproceedings{holder2016road,
	title={From on-road to off: transfer learning within a deep convolutional neural network for segmentation and classification of off-road scenes},
	author={Holder, Christopher J and Breckon, Toby P and Wei, Xiong},
	booktitle={European Conference on Computer Vision},
	pages={149--162},
	year={2016},
	organization={Springer}
}

@article{tian2019contrastive,
	title={Contrastive multiview coding},
	author={Tian, Yonglong and Krishnan, Dilip and Isola, Phillip},
	journal={arXiv preprint arXiv:1906.05849},
	year={2019}
}

@article{zhao2020contrastive,
	title={Contrastive Learning for Label-Efficient Semantic Segmentation},
	author={Zhao, Xiangyun and Vemulapalli, Raviteja and Mansfield, Philip and Gong, Boqing and Green, Bradley and Shapira, Lior and Wu, Ying},
	journal={arXiv preprint arXiv:2012.06985},
	year={2020}
}
	
	%
	%

\end{document}